\documentclass[conference]{IEEEtran}
\IEEEoverridecommandlockouts
\usepackage{cite}
\usepackage{amsmath,amssymb,amsfonts}
\usepackage{algorithmic}
\usepackage{graphicx}
\usepackage{textcomp}
\usepackage{xcolor}
\usepackage{float}
\usepackage{stfloats}
\usepackage{colortbl}
\usepackage{makecell}
\usepackage{hyperref}
\usepackage{cleveref}
\usepackage{fontawesome5}
\usepackage{xcolor}
\usepackage{xspace}
\usepackage{caption}
\usepackage{booktabs}
\newcommand{\eg}{\emph{e.g.}\xspace}
\newcommand{\ie}{\emph{i.e.}\xspace}

\def\BibTeX{{\rm B\kern-.05em{\sc i\kern-.025em b}\kern-.08em
    T\kern-.1667em\lower.7ex\hbox{E}\kern-.125emX}}
\begin{document}

\title{FDDet: Achieving Data-Efficient Food Defect Detection Under Real-World Scenarios}

\author{\IEEEauthorblockN{Ruihao Xu, Yong Liu, Yansong Tang$^\dagger$}
\IEEEauthorblockA{\textit{Tsinghua Shenzhen International Graduate School, Tsinghua University}}}

\twocolumn[{%
	\renewcommand
	\twocolumn[1][]{#1}%
	\maketitle
	\begin{center}
        \centering
        \vspace{-15pt}
        \includegraphics[width=0.85\textwidth]{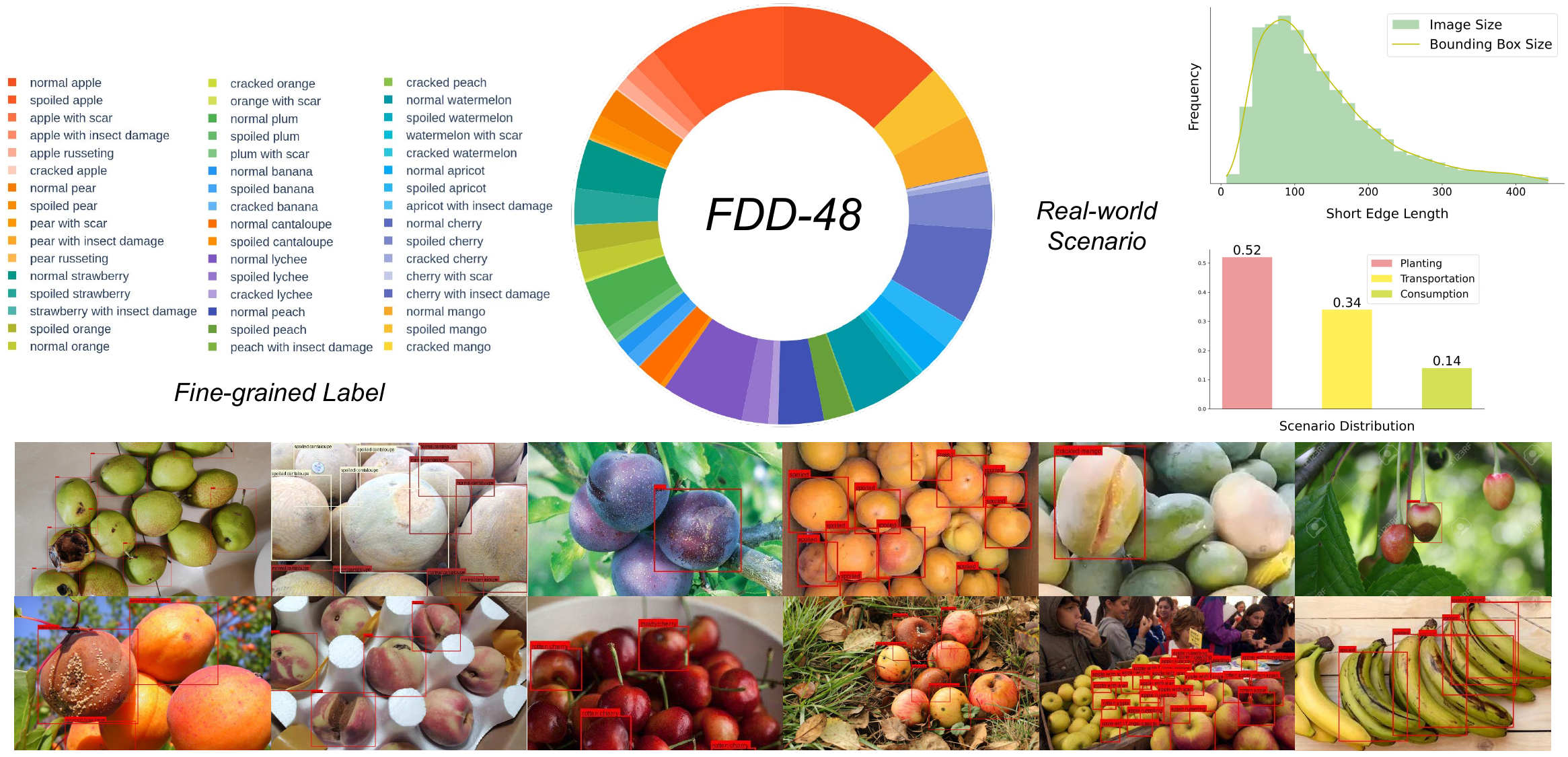}
        \captionof{figure}{FDD-48: A food defect detection dataset with fine-grained annotations to closely align with practical application scenarios. It encompasses 13 food types and 48 defect types, covers diverse scenarios like Planting, Transportation, and Consumption, and ensures balanced distributions of image sizes and bounding box sizes for real-world applications.}
        \label{fig:dataset}
	\end{center}
}]

\begingroup
\renewcommand{\thefootnote}{}
\footnotetext{\hrule\vspace{5pt}$^\dagger$Corresponding author: tang.yansong@sz.tsinghua.edu.cn.\\This work was supported in part by the National Key Research and Development Program of China under Grant 2023YFF1105101.}
\endgroup

\begin{abstract}
Food defect detection is critical for automated quality control, yet existing studies lack unified benchmarks and suffer from data scarcity. We introduce \textbf{FDD-48}, a comprehensive dataset with fine-grained annotations across 13 food types and 48 defect categories under diverse real-world conditions. To improve detection with limited labeled data, we propose \textbf{FDDet}, a semi-supervised framework featuring two key components: (1) \textbf{BBoxMixUp}, a data augmentation technique that mixes same-category defect regions to reduce spurious feature associations, and (2) \textbf{CGPC} (\textbf{C}onsistency-\textbf{G}uided \textbf{P}seudo-Label \textbf{C}alibration), which filters pseudo-labels based on intra-sample consistency. Experiments show FDDet significantly outperforms mainstream detectors on FDD-48, demonstrating its effectiveness for food defect detection under data-limited scenarios.
\end{abstract}

\section{Introduction}

Food safety and quality assurance are core concerns in the global food industry. Traditional detection methods rely on manual inspection or basic image processing \cite{birdnest, chen2021machine, patel2021monochrome}, while machine learning approaches \cite{zhang2021identification, thanasarn2019automated, mandal2022color, bhargava2023machine} improved accuracy but remain limited by training data quality. Deep learning has enabled automated, real-time food defect detection \cite{fu2021fast, mohana2022novel, chen2023intelligent, yang2024lightweight}, yet most studies rely on proprietary datasets from controlled environments with homogeneous backgrounds and binary classification labels, lacking fine-grained defect categories and localization annotations. To address these gaps, we introduce the \textbf{FDD-48} dataset. As shown in \cref{dataset comparison}, FDD-48 encompasses multiple fine-grained food defect types, instance-level annotations, and diverse real-world backgrounds. We deliberately maintained a moderate dataset size to simulate real-world data scarcity, enabling effective evaluation of model generalization and practical robustness.

We subsequently evaluated the performance of several mainstream general-purpose object detection models on FDD-48, with results presented in \cref{tab:main_result}. The models exhibited suboptimal performance, which we attribute to overfitting caused by limited annotated data, further highlighting the limitations of general-purpose models in data-constrained scenarios. To investigate further, we analyzed FDD-48 samples and visualized model outputs. We observed substantial variation in non-defect features (e.g., shape, color) across defect types, which the limited training set failed to cover comprehensively. Moreover, food defects often occupy small surface areas, providing insufficient supervisory signals. These factors collectively predispose models to overfit to non-defect features.

\begin{table}[!t]
\caption{Comparison with other public food defect datasets.}
\centering
\setlength{\tabcolsep}{0.08cm}
% \footnotesize
\small
\begin{tabular}{lccccccc}
\toprule
\textbf{Dataset} & \textbf{\makecell{Diverse \\ Scenarios}} & \textbf{\makecell{Label \\ Level}} & \textbf{Cls.} & \textbf{\makecell{Real \\ Images}} & \textbf{Labels} & \textbf{\makecell{Avg. \\ Inst. \\ /Img}} \\ 
\midrule
FruitNet\cite{meshram2022fruitnet} & \textcolor{red}{\faTimesCircle} & Global & 18 & 14700 & 14700 & $\approx3$ \\ 
HDFruits\cite{visapp23} & \textcolor{red}{\faTimesCircle} & Global & 4 & 10000 & 10000 & 1 \\ 
MeatQA\cite{ulucan2019meat} & \textcolor{red}{\faTimesCircle} & Global & 2 & 948 & 948 & 1 \\ 
PotatoDS\cite{arshaghi2023potato} & \textcolor{red}{\faTimesCircle} & Global & 5 & 4813 & 4813 & 1 \\ 
Fmaaa307\cite{fmaaa307_2023} & \textcolor{red}{\faTimesCircle} & Instance & 4 & 1200 & 1200 & 1 \\ 
VisA-Food\cite{zou2022spot} & \textcolor{red}{\faTimesCircle} & Instance & 43 & 4603 & 4603 & $\approx2$ \\
\textbf{FDD-48 (ours)} & \textcolor[HTML]{00CC00}{\faCheckSquare} & \textbf{Instance} & \textbf{48} & \textbf{4000} & \textbf{15913} & $\mathbf{\approx11}$ \\ 
\bottomrule
\end{tabular}
% \vspace{-5pt}
\label{dataset comparison}
\end{table}

To address these challenges, we first propose \textbf{BBoxMixUp}, a novel data augmentation technique tailored for food defect detection. Unlike traditional mixing methods \cite{zhang2017mixup, yun2019cutmix, hendrycks2019augmix, hendrycks2022pixmix, huang2023ipmix}, which blend images and labels globally, BBoxMixUp performs localized mixing exclusively within bounding boxes of the same defect category. This approach enhances diversity in non-defect features while disrupting spurious correlations between defect types and non-defect features caused by data scarcity, thereby mitigating overfitting.

Recognizing that unlabeled data is often more accessible than labeled data in real-world applications, we further explore semi-supervised learning (SSL) paradigms for food defect detection to leverage the 2,497 unlabeled samples in FDD-48. However, directly applying common semi-supervised object detection frameworks \cite{liu2021unbiased} led to training collapse. We attribute this to two factors: (1) unupdated buffer weights (e.g., Batch Normalization parameters) prevented the teacher model from adapting to domain shifts between labeled and unlabeled data; (2) conventional high-confidence pseudo-label thresholds (e.g., 0.9) were incompatible with FDD-48’s class similarity and generally low model confidence, discarding substantial valid supervisory signals. To resolve these issues, we implemented some modifications to \cite{liu2021unbiased}, successfully stabilized training and significantly improved model performance.

To further refine and calibrate pseudo-label quality, we propose \textbf{CGPC} (\textbf{C}onsistency-\textbf{G}uided \textbf{P}seudo-Label \textbf{C}alibration). CGPC optimizes initial teacher-generated pseudo-labels by enforcing multi-dimensional consistency constraints: (1) \textit{Context-Semantic Consistency} leverages the prior that objects within an image typically belong to the same high-level semantic category (e.g., food type), unifying all pseudo-labels in a single image to the most frequent food type; (2) \textit{Visual-Semantic Consistency} ensures visually similar regions receive consistent labels based on features extracted by an external pretrained backbone; (3) \textit{Spatial Consistency} removes spatially redundant pseudo-labels via non-maximum suppression (NMS)-like methods. Through CGPC’s multi-dimensional calibration, we provide higher-quality supervisory signals for the student model, further enhancing the effectiveness of semi-supervised learning.

The main contributions of this paper are as follows:

\begin{itemize}
    \item We introduce FDD-48, a novel benchmark dataset for food defect detection featuring diverse real-world scenarios, fine-grained defect categories, and instance-level annotations.
    \item We propose BBoxMixUp, a data augmentation technique tailored for food defect detection, which performs localized mixing to enhance non-defect feature diversity and mitigate overfitting under data scarcity.
    \item We enable effective semi-supervised learning (SSL) on FDD-48 by adapting existing frameworks for stable training,  and introduce CGPC, a novel method that refines pseudo-labels through multi-dimensional consistency constraints to further improve SSL`s performance.
    \item Through extensive experiments, we demonstrate the limitations of existing models on FDD-48 and validate the significant performance gains achieved by our methods.
\end{itemize}
\section{FDD-48: A Food Defect Detection Dataset}

Data is a vital driver of deep learning, making task-specific datasets indispensable. Unlike general object detection datasets, food defect detection datasets require annotators with specific expertise, making large-scale annotation difficult. As a result, existing research datasets are limited in terms of sample size, diversity of data collection scenarios, number of defect types, and granularity of defect types, hindering their real-world applications.

\subsection{Food Objects and Defect Types}

We initiated our study with fruits due to their high consumption rates and significant economic value. Fruits are highly perishable and susceptible to a variety of defects during planting, production, transportation, and consumption stages. Monitoring and detecting these defects are crucial for ensuring food quality and safety. We selected 13 fruit types that are among the most consumed on a global scale: apple, apricot, banana, cantaloupe, cherry, lychee, mango, orange, peach, pear, plum, strawberry, and watermelon. For each fruit, we identified the most common defect types occurring in different scenarios. In total, we identified 48 distinct food defect types, as shown in Figure~\ref{fig:dataset}.

\subsection{Image Acquisition and Preprocessing}

We used the Selenium Python library to automatically crawl images from online search engines, employing the previously defined "food category + defect type" as keywords, resulting in a total of 50,000 raw images. Given that web-crawled images often contain irrelevant or noisy data, manual filtering was impractical due to the large volume. To address this, we employed the multimodal large model MiniCPM-V-2.6~\cite{yao2024minicpm} to automatically assess whether the images contained the target food type and removed those without it. Subsequently, to eliminate duplicate images, we used DINOv2~\cite{oquab2023dinov2} to extract image features, calculated feature similarity, and removed images with similarity exceeding a preset threshold. Finally, through manual screening, we obtained 4,000 images with diverse scenes and balanced food categories as the final dataset.

\subsection{Dataset Annotation}
For initial annotating with bounding boxes, we leveraged YOLO-World~\cite{cheng2024yolo}, an open-source object detector capable of utilizing textual prompts to detect corresponding objects. By inputting a vocabulary list of the 13 fruit names as a prompt, YOLO-World generated bounding box annotations for both normal and defective fruits within the images. 
Subsequently, we conducted manual verification to annotate any missed fruit instances and to assign specific defect labels to all bounding boxes, ensuring that the dataset had high-quality annotations suitable for training and evaluating defect detection models.

\subsection{Statistical Characteristics of the Dataset}

The dataset consists of 4,000 images, including 1,503 labeled images and 2,497 unlabeled images. The labeled data contains 15,913 fruit instances, with an average of approximately 10.58 objects per image. The distribution of images across 13 fruit categories is balanced, of which 39\% include defective fruits. The dataset covers 48 defect types, with their frequency distribution presented in a pie chart, alongside statistical distributions of image and bounding box sizes. Additionally, the dataset encompasses diverse scenarios, including planting, transportation, and consumption. Finally, the set of labeled data is randomly split into training and testing sets in a 7:3 ratio.

\section{FDDet}
In this section, we propose FDDet, a food defect detection model based on RTMDet. FDDet incorporates three key components: (1) BBoxMixUp, a localized data mixing strategy for food defect data; (2) adapted semi-supervised learning to address training collapse; and (3) CGPC, a consistency-based pseudo-label calibration strategy.

\begin{figure}[t]
    \centering
    \includegraphics[width=0.5\textwidth]{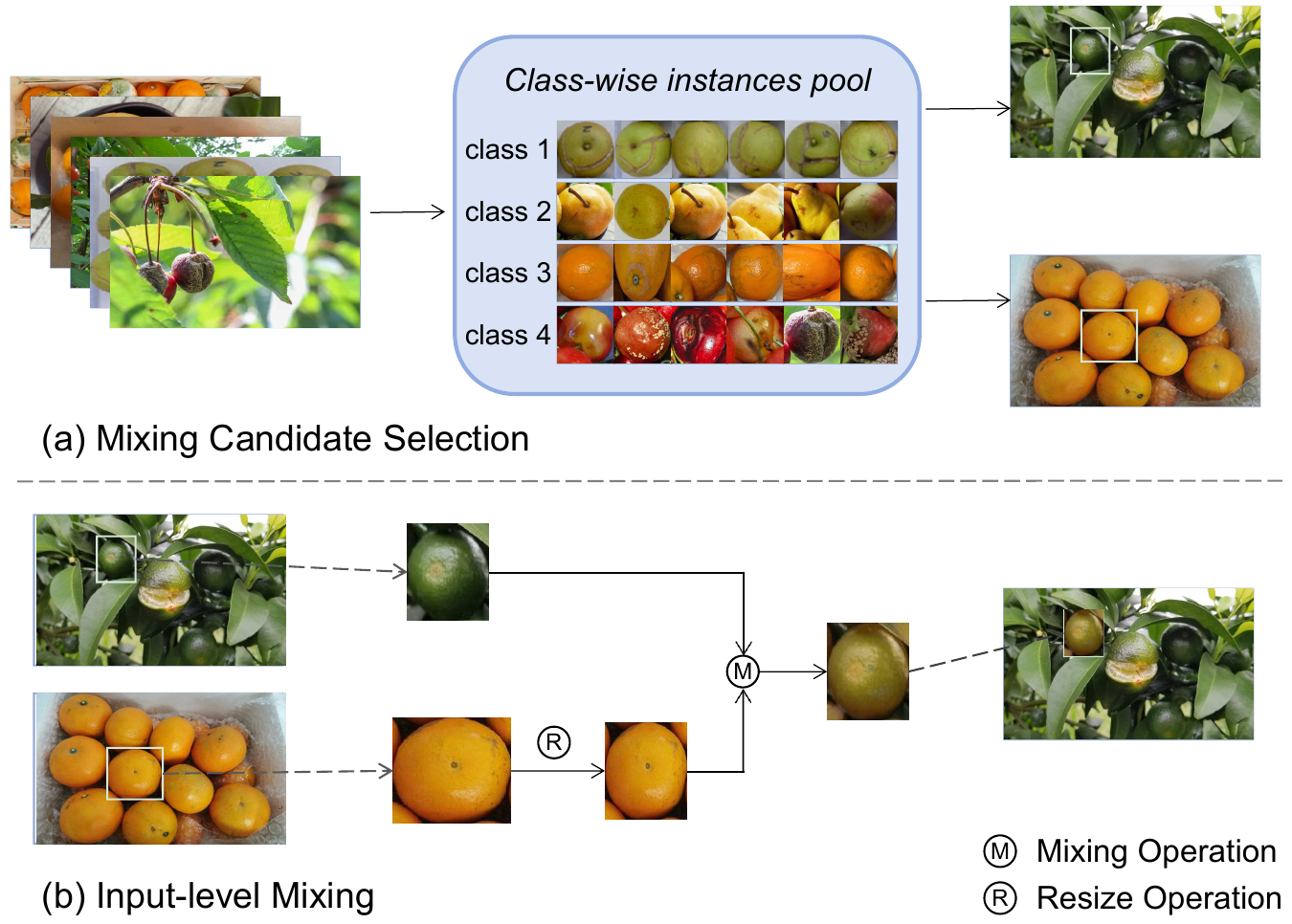}  % 这里替换为你的图片路径
    \caption{Illustration of BBoxMixUp. \textbf{(a) Mixing Candidate Selection.} Bounding boxes of the same class are pooled and randomly paired for mixing. \textbf{(b) Localized Mixing.} Regions from different images are blended to create new instance.}
    \label{fig:mix}
\end{figure}

\subsection{BBoxMixUp}
Food defect detection faces three key challenges: (1) defect regions are often very small, providing insufficient supervisory signals; (2) limited training data causes models to overfit to sample appearances rather than defect features; and (3) traditional augmentation methods (\eg MixUp and CutMix) focus on global transformations, neglecting localized defect characteristics.

To address these issues, we propose \textbf{BBoxMixUp}, a localized data mixing strategy that blends object regions of the same class from different images. As shown in Figure~\ref{fig:mix}, BBoxMixUp consists of two stages:
\textbf{(1) Mixing Candidate Selection.} For each target bounding box, we randomly select a candidate box from a different image containing an object of the same class.
\textbf{(2) Mixing Operation.} The candidate region is resized to match the target region dimensions. A mixing ratio $\lambda$ is sampled from $\text{Beta}(\alpha, \beta)$, and the target region is replaced by a weighted combination of both regions.

\begin{figure}[]
    \centering
    \includegraphics[width=\columnwidth]{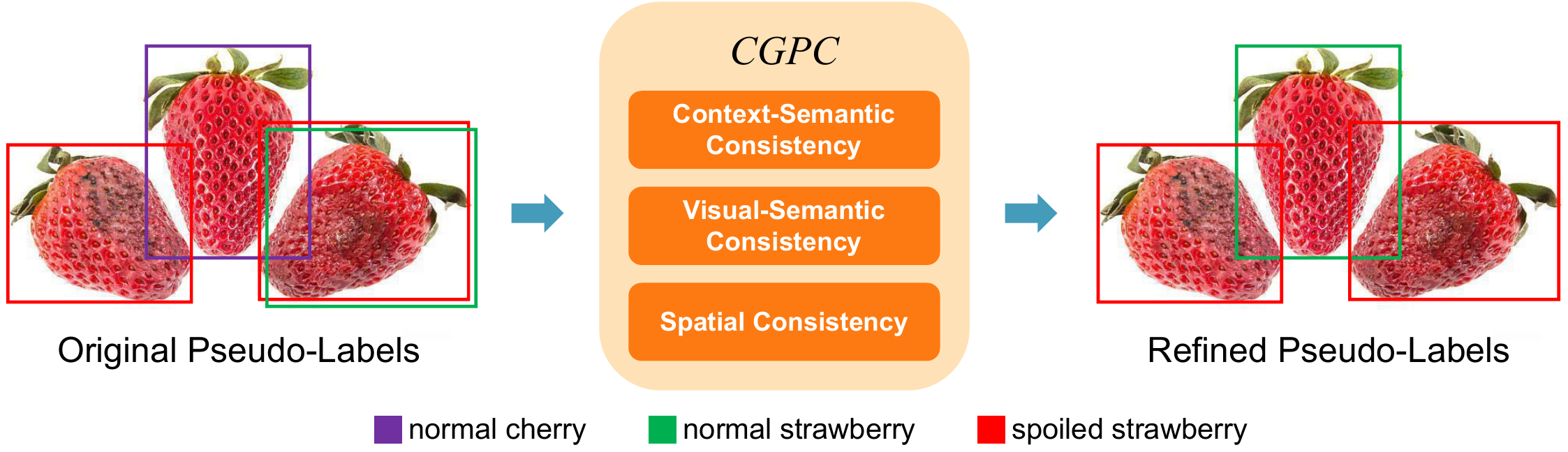}
    \caption{CGPC applies multi-dimensional consistency calibrations to initial, noisy pseudo-labels from a teacher model, producing higher-quality pseudo-labels for student model training.}
    \label{fig:cgpc}
\end{figure}

\subsection{Semi-Supervised Learning for Food Defect Detection}
To leverage the 2,497 unlabeled samples in FDD-48, we adopt the semi-supervised object detection framework~\cite{liu2021unbiased}, which employs a teacher-student paradigm where the teacher generates pseudo-labels for unlabeled data and is updated via exponential moving average (EMA) from the student. However, direct application causes training collapse—the teacher model suddenly produces empty predictions after a few iterations. We identify two causes and propose corresponding solutions:

\textbf{(1) EMA Buffer Update.} The original framework does not update the teacher's buffer weights (\eg, BatchNorm statistics) via EMA. For small datasets like FDD-48 with potential domain shifts between labeled and unlabeled data, this causes the teacher to produce erroneous pseudo-labels. We enable EMA updates for buffer weights to allow dynamic adaptation.

\textbf{(2) Lower Confidence Threshold.} Standard high thresholds (\eg, 0.9) discard many reliable pseudo-labels, as high inter-class similarity in food defects leads to inherently lower confidence scores. We use a lower threshold (0.35) to retain more valuable pseudo-labels.

\begin{table*}[!t]
\centering
\caption{Comparison with State-of-the-Art object detectors on FDD-48. The best results for each metric are \textbf{bolded}, and the second-best results are \underline{underlined}. Our method achieves new state-of-the-art results.}
\begin{tabular}{l|ccc|ccc}
\toprule
\textbf{Method} & \textbf{Backbone} & \textbf{Param.(M)} & \textbf{Pre-trained Data} & $\mathbf{mAP_{50:95}(\%)}$ & $\mathbf{mAP_{50}(\%)}$ & $\mathbf{mAP_{75}(\%)}$ \\ 
\midrule
YOLOX~\cite{ge2021yolox} & YOLOX-x & 99.0 & COCO & 38.9 & 44.3 & 42.7\\
RTMDet~\cite{lyu2022rtmdet} & RTMDet-x & 94.8 & COCO & 40.0 & 45.6 & 43.7 \\
DAB-DETR~\cite{liudab} & R-50 & 43.7 & COCO & 34.7 & 40.2 & 38.3 \\
DINO~\cite{zhangdino} & R-50 & 47.6 & COCO & 38.2 & 42.8 & 41.0 \\
GLIP~\cite{li2022glip} & Swin-T & 231.8 & O365, GoldG, CC3M, SBU & 25.0 & 27.7 & 26.6 \\
DDQ~\cite{zhang2023ddq} & R-50 & 48.5 & COCO & 37.6 & 41.6 & 40.4 \\
EfficientDet~\cite{tan2020efficientdet} & EfficientNet-b3 & 12.0 & COCO & \underline{41.7} & 46.9 & \underline{45.0} \\
VitDet~\cite{li2022vitdet} & ViT-B & 108.1 & COCO & 41.5 & \underline{47.0} & 44.4 \\
CO-DETR~\cite{zong2022codetrs} & Swin-L & 235.2 & Objects365, COCO & 40.0 & 43.7 & 42.8 \\ 
YOLOv10~\cite{wang2024yolov10} & YOLOv10-x & 29.5 & COCO & 41.2 & 46.3 & \underline{45.0} \\ 
\rowcolor[gray]{0.9}FDDet (ours) & RTMDet-x & 94.8 & COCO & \textbf{42.2} & \textbf{49.0} & \textbf{46.0}  \\
\bottomrule
\end{tabular}
\label{tab:main_result}
\end{table*}

\begin{table}[]
\caption{Component Analysis. BM, SSL, and CGPC denote BBoxMixUp, semi-supervised learning, and Consistency-Guided Pseudo-Label Calibration, respectively.}
\setlength{\tabcolsep}{3pt}
\begin{tabular}{ccc|ccc}
\toprule
\textbf{BM} & \textbf{SSL} & \textbf{CGPC} & $\mathbf{mAP_{50:95}(\%)}$ & $\mathbf{mAP_{50}(\%)}$ & $\mathbf{mAP_{75}(\%)}$ \\
\midrule
 &  &  & 40.0 & 45.6 & 43.7 \\
\checkmark &  &  & 41.4 & 47.9 & 44.7 \\
\checkmark & \checkmark &  & 41.7 & 48.3 & 45.2 \\
\checkmark & \checkmark & \checkmark & \textbf{42.2} & \textbf{49.0} & \textbf{46.0} \\
\bottomrule
\end{tabular}
\label{tab:component}
\end{table}

\subsection{CGPC}
To further enhance the effectiveness of semi-supervised learning, we analyzed the characteristics of the teacher model's predictions on unlabeled data. We observed that when using a relatively low threshold for initial pseudo-labels filtering, the teacher model often outputs multiple pseudo-labels with different categories but highly overlapping bounding boxes, among which the correct labels are frequently included (shown in Figure~\ref{fig:cgpc}). Therefore, we proposed the CGPC (Consistency-Guided Pseudo-Label Calibration) strategy to identify the correct pseudo-labels after initial pseudo-labels filtering. The core idea of CGPC lies in refining and calibrating the initial pseudo-labels generated by the teacher model by enforcing "consistency" across multiple dimensions and levels. Here, "consistency" is reflected in the following three aspects:

\textbf{Context-Semantic Consistency.} Since we collect data through web retrieval, in most cases, food instances within the same image belong to the same food category—for example, both fresh and spoiled apples are apples. Therefore, different pseudo-labels within the same image should align at the high-level semantic level of food category. To achieve this consistency, we adopted a straightforward method: we first identify the most frequently occurring food category among all pseudo-labels in a single image, then replace the food category of all pseudo-labels with this most frequent one.

\textbf{Visual-Semantic Consistency.} Visually similar regions likely belong to the same category. CGPC extracts region features using a pre-trained visual backbone (e.g., RegNet~\cite{radosavovic2020regnet}) and computes pairwise similarities. For each pseudo-label, regions with similarity above a threshold are identified as peers, and the pseudo-label is replaced by the most frequent one among them.

\textbf{Spatial Consistency.} For bounding boxes with high spatial overlap, they are highly likely to point to the same object instance, so their pseudo-labels should be unified. After pseudo-label correction based on context-semantic or visual-semantic consistency, some regions may still have multiple pseudo-labels of the same category. To address this, CGPC employs an IOU-based method similar to NMS after each of the two correction processes to remove duplicate pseudo-labels, eliminating redundant detections in space.

% \subsection{Implementation Details}
% To ensure training stability while reducing training time overhead, FDDet employs a two-stage training scheme: first conducting fully supervised training until model convergence, then initiating semi-supervised training from the best-performing checkpoint.
% For BBoxMixUp, the default Beta distribution parameters are $\alpha=1$ and $\beta=0.5$. In the semi-supervised learning framework, the default threshold for initial pseudo-label filtering is 0.35. For CGPC, RegNet is used as the default pre-trained visual backbone. Both training stages use a batch size of 32. The first training stage runs for 300 epochs. The second stage runs for 1000 iterations, with each batch comprising a 1:1 ratio of labeled to unlabeled data. We use the AdamW optimizer with a learning rate of 0.004.
\section{Experiment}

\subsection{Experiment Settings}
We selected some state-of-the-art models for general object detection as baselines. These models were first initialized with pre-trained weights then fine-tuned on the FDD-48 training set, and their performance was evaluated on the test set.
Model performance was evaluated using mean Average Precision (mAP). We report mAP@0.5, mAP@0.75, and mAP@0.5:0.95.

\subsection{Results on FDD-48}
On the FDD-48 dataset, our FDDet achieves new state-of-the-art performance with 42.2\% $\text{mAP}_{\text{50:95}}$, 49.0\% $\text{mAP}_{\text{50}}$, and 46.0\% $\text{mAP}_{\text{75}}$. This surpasses prior top scores by 0.5\%, 2.0\%, and 1.0\% in $\text{mAP}_{\text{50:95}}$, $\text{mAP}_{\text{50}}$, and $\text{mAP}_{\text{75}}$ respectively. Furthermore, FDDet substantially outperforms its baseline RTMDet by 2.2\%, 3.4\%, and 2.3\% in $\text{mAP}_{\text{50:95}}$, $\text{mAP}_{\text{50}}$, and $\text{mAP}_{\text{75}}$ respectively, highlighting its superiority and the effectiveness of our proposed modifications.

\subsection{Component Analysis.}
Results shown in the Table~\ref{tab:component} demonstrate the incremental benefits on FDD-48 test set of each component proposed in this paper. BBoxMixUp (BM) alone improved mAP$_{50:95}$/mAP$_{50}$/mAP$_{75}$ by +1.4\%/+2.3\%/+1.0\% over the baseline (\ie RTMDet). Adding semi-supervised learning (SSL) to BM further yielded +0.3\%/+0.4\%/+0.5\% gains. Finally, Consistency-Guided Pseudo-Label Calibration (CGPC) built upon BM+SSL, contributing an additional +0.5\%/+0.7\%/+0.8\%, leading to the full model's 42.2\% mAP$_{50:95}$, 49.0\% mAP$_{50}$ and 46.0\% mAP$_{75}$.

\section{Conclusion}
This paper introduces FDD-48, a fine-grained food defect detection benchmark with diverse real-world scenarios, and proposes FDDet, a semi-supervised framework integrating BBoxMixUp for localized augmentation and CGPC for consistency-guided pseudo-label calibration. Experiments demonstrate FDDet achieves state-of-the-art performance on FDD-48, validating the effectiveness of our approach for food defect detection under limited annotation.

\bibliographystyle{IEEEbib}
\bibliography{main}

\newpage
\appendix

\section{More Dataset Samples}
\label{appendix:datasetmore}

\begin{figure*}[!htbp]
    \centering
    \includegraphics[width=\textwidth]{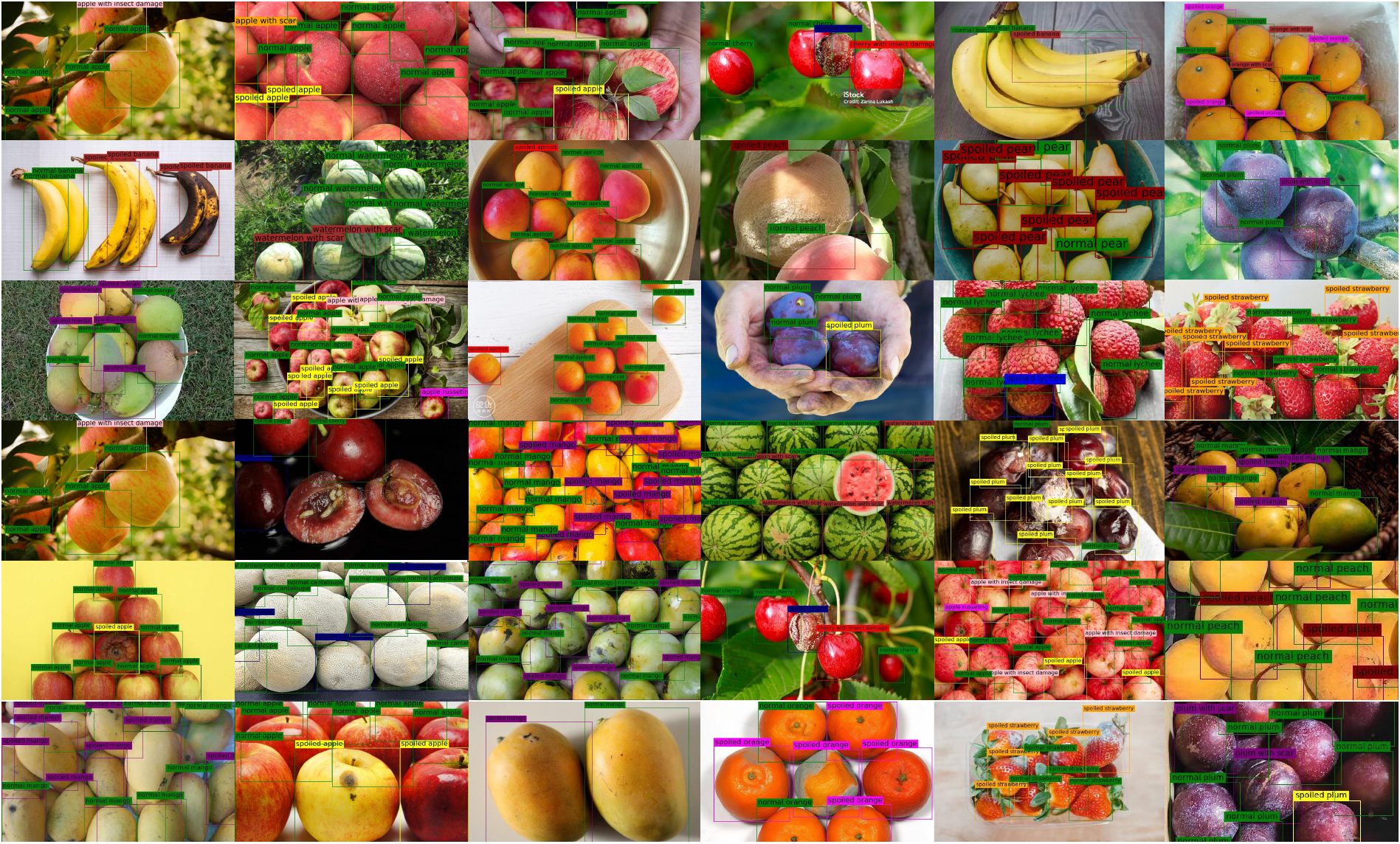}
    \caption{More sample images from FDD-48, demonstrating the diversity of food types and defect categories.}
    \label{fig:datasetmore}
\end{figure*}

We provide additional sample images from FDD-48 to illustrate the diversity of food types and defect appearances in our dataset. As shown in \cref{fig:datasetmore}, the dataset covers a wide range of real-world food defect scenarios.

\section{Defect Category Illustrations}
\label{appendix:datasetinfo}

\begin{figure*}[!htbp]
    \centering
    \includegraphics[width=\textwidth]{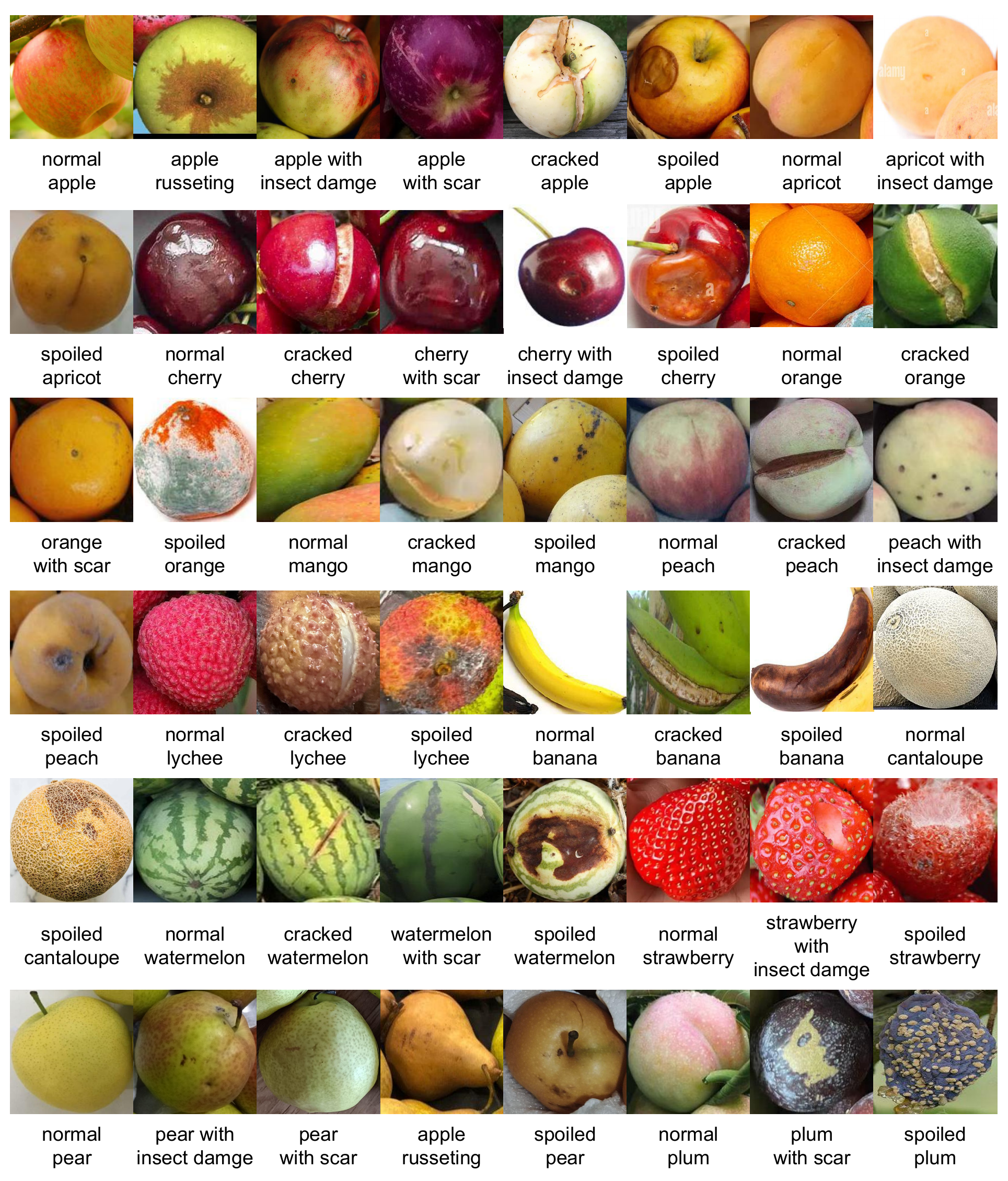}
    \caption{Illustrative examples for each defect category in FDD-48, showing the visual characteristics of different food defects.}
    \label{fig:datasetinfo}
\end{figure*}

To better understand the characteristics of each defect type, we present illustrative examples for individual defect categories in \cref{fig:datasetinfo}. Each example highlights the visual appearance and distinguishing features of the corresponding defect.

\end{document}